%% file: main.tex
\documentclass[sigconf]{acmart}

\usepackage{pgfplots} 
\usepackage{enumitem}
\usepackage{subcaption}
\usepackage{url}
\usepackage{hyperref}
\AtBeginDocument{%
  \providecommand\BibTeX{{%
    \normalfont B\kern-0.5em{\scshape i\kern-0.25em b}\kern-0.8em\TeX}}}

\setcopyright{rightsretained}
\copyrightyear{2022}

\acmConference[]{}{}{}

\begin{document}

\title{Study of detecting behavioral signatures within DeepFake videos}

\author{Qiaomu Miao}
\affiliation{%
  \institution{Stony Brook University}
   \country{USA}
}
\email{qiamiao@cs.stonybrook.edu}

\author{Sinhwa Kang}
\affiliation{%
  \institution{}
   \country{USA}
  }
\email{sinhwa.kang@gmail.com}

\author{Stacy Marsella}
\affiliation{
  \institution{Northeastern University}
   \country{USA}
}
\email{stacymarsella@gmail.com}

\author{Steve DiPaola}
\affiliation{%
 \institution{Simon Fraser University}
 \country{Canada}
 }
 \email{sdipaola@sfu.edu}

\author{Chao Wang}
\affiliation{
 \country{USA}
 }
\email{chaowang15@gmail.com}

\author{Ari Shapiro}
\affiliation{
 \country{USA}
 }
\email{ariyshapiro@gmail.com}

\renewcommand{\shortauthors}{Miao et al.}

\begin{abstract}
There is strong interest in the  generation of synthetic video imagery of people talking for various purposes, including entertainment, communication, training, and advertisement.  With the development of deep fake generation models, synthetic video imagery will soon be visually indistinguishable to the naked eye from a naturally capture video. In addition, many methods are continuing to improve to avoid more careful, forensic visual analysis. Some deep fake videos are produced through the use of facial puppetry, which directly controls the head and face of the synthetic image through the movements of the actor, allow the actor to 'puppet' the image of another. In this paper, we address the question of whether one person’s movements can be distinguished from the original speaker by controlling the visual appearance of the speaker but transferring the behavior signals from another source. We conduct a study by comparing synthetic imagery that: 1) originates from a different person speaking a different utterance, 2) originates from the same person speaking a different utterance, and 3) originates from a different person speaking the same utterance. Our study shows that synthetic videos in all three cases are seen as less real and less engaging than the original source video. Our results indicate that there could be a behavioral signature that is detectable from a person's movements that is separate from their visual appearance, and that this behavioral signature could be used to distinguish a deep fake from a properly captured video.
\end{abstract}




\begin{teaserfigure}
  \includegraphics[width=1\textwidth]{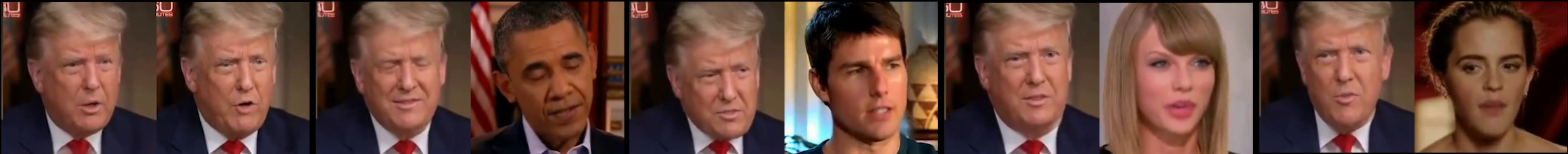}
  \caption{Driving a facial performance by different sources. Lip syncing to voice is consistent, but other nonverbals such as eye movements, head movements and facial expressions are copied from the original actor. The videos are shown in pairs; the left video of the pair shows the results of the DeepFake, the right video shows the driving actor. }
 \label{fig:teaser}
\end{teaserfigure}

\maketitle

\input{introduction}

\input{relatedwork}
\input{studydesign}

\input{results}
\input{discussion}

\bibliographystyle{ACM-Reference-Format}

\bibliography{studydeepfakevideo}


\end{document}

%% file: introduction.tex
\section{Introduction}
Recent advances have allowed the generation of synthetic imagery of human faces. Such imagery can be synthesized as a a static, unmoving image like a photograph, as well as a moving image, like a video recording of a real person. In some cases, the human-like imagery is able to surpass the Uncanny Valley \cite{mori1970bukimi} to casual viewers who haven't done a thorough forensic investigation of the media. In addition, advances in facial puppetry have allowed speaking performances of those synthetic heads and faces that can mimic the speaking movements of the controlling actor.

There are many beneficial uses of such synthetic imagery including entertainment, training, health care, and automated interactions. There are also numerous nefarious uses of such synthetic imagery, such as political misinformation, falsified video or photographic content. As such, there is an interest in determining whether or not such synthetic imagery is a representation of the real person, or of a synthetic image. Thus, efforts have also looked at detection such synthetic imagery, or 'DeepFake' detection.

Most of the recent attempts in DeepFake detection have been focused on visual artifacts in synthetic images/videos, such as improper illumination, missing details in facial regions, facial warping or temporal inconsistencies between frames \cite{matern2019exploiting, nirkin2020deepfake, li2019exposing, sabir2019recurrent}. However, with the rapid development of DeepFake generation methods and large improvement in visual quality of synthetic output, these kind of methods will undoubtedly become less effective and may even fail to correctly identify the fake images/videos, as the state-of-the-art generative models can already synthesize high-resolution images which are almost indistinguishable by humans \cite{karras2019style, karras2020analyzing}. 

In video synthesis, great interest has been placed in the generation of people talking for the purpose of fast and flexible content creation, with potential uses such as video conferencing and other remote communications. Like synthetic images, it is only a matter of time before synthetic talking videos become indistinguishable from real ones in visual appearances, limiting the value of artifacts-based DeepFake detection methods in the future. However, for human to identify a talking video as real, a realistic visual appearance is only the most basic requirement. Humans also judge the naturalness of the person's talking behavior, by comparing the speaking style of the person in video with their impressions, and checking the correspondence between the talking behavior and utterance. For example, the well-known Tom Cruise DeepFakes are convincing not only because the deep fake looks like Tom Cruise, but acts like him (or at least acts like a satire of him) as well with the behaviors performed by the impersonator who knows him very well. However, state-of-the-art generative models usually ignore the behavioral signatures of a person in video generation and it is the same with the case of DeepFake detection methods.  
With this in mind, we raise the following questions that bring up our study: Could DeepFakes be detected by understanding a person's behavioral 'signature' then comparing that signature to a video of that person? Could a person's speaking style be transferred onto a digital image of that person and become a convincing representation of that person?

Recent facial puppetry algorithms have enabled us to explore these questions, which have been shown to be able to transfer a persons head, eye and lip movements onto an image of another person, allowing one person to ‘puppet’ the image of another. In this paper, we address the question of whether movements transferred from another person can be distinguished while keeping the visual appearance of the original speaker in a talking video. We conduct a study and compare synthetic imagery that: 1) originates from a different person speaking a different utterance, 2) originates from the same person speaking a different utterance, and 3) originates from a different person speaking the same utterance. Our study shows that synthetic videos in all three cases are seen as less real and less engaging than the original source video.

%% file: relatedwork.tex
\section{Related Work}

\subsection{DeepFake Generation Methods}

DeepFake generation methods can be generally be divided in to 4 categories: whole face synthesis, attribute manipulation, identity swap, and expression swap.

The first two categories mainly generates static face images. Whole face synthesis methods generate entirely non-existent face images, mostly using the powerful generative adversarial networks (GANs). The current state-of-the-art GANs, such as PGGAN \cite{karras2017progressive}, StyleGAN \cite{karras2019style}, StyleGAN2 \cite{karras2020analyzing} are able to generate very realistic high-resolution human faces. Attribute manipulation methods create fake images by modifying certain attributes of the face, such as hair, skin, gender, adding eyeglasses, etc. Most of these kind of methods edit the latent space of the input images to generate images with modified facial attribute. The state-of-the-art models such as StarGAN \cite{choi2018stargan} STGAN \cite{liu2019stgan}, HifaFace \cite{gao2021high}, and TediGAN \cite{xia2021tedigan} has enabled high fidelity image creation, or even enable interactive facial attribute editing.

Identity swap methods generate fake videos by replacing the face of the original person in a video with the face of another person. This kind of methods include classical computer graphics based methods such as FaceSwap \footnote{\url{https://github.com/
MarekKowalski/FaceSwap}} and deep learning based methods like DeepFakes \footnote{\url{https://github.com/
deepfakes/faceswap}}. The recent deep learning based methods including FaceShifter \cite{li2020advancing}, InfoSwap \cite{gao2021information}, and MegaFS \cite{zhu2021one} have enabled generating face swapped videos of high fidelity, or with few samples required in training. However, this kind of methods just replace the face region of the target person with the source person, while keeping other visual information, e.g. hair, clothes, background unchanged.  

Expression swap methods aim to create fake videos by replacing the expression of person in video with that of another person, which is also known as face reenactment. Expression swaps methods can also be divided into computer graphics-based methods, e.g., Face2Face \cite{thies2016face2face} and deep-learning based methods, such as NeuralTextures \cite{thies2019deferred}, First Order Motion Model (FOMM) \cite{siarohin2019first}, and HeadGAN \cite{doukas2021headgan}. These methods shows generally good results, but still showing artifacts in certain cases. In addition, there are also other highly related works that only modify the mouth region to correspond with the input audio, known as lipsyncing. Current lipsyncing models can generate very realisitic lip synced videos \cite{suwajanakorn2017synthesizing, prajwal2020lip, lahiri2021lipsync3d}.  

\subsection{DeepFake Detection Methods}
The development of these powerful DeepFake generation methods are beneficial to various applications, 
 but also pose greater risks if adopted for malicious uses. Therefore, great efforts have been made by the research community to identify the images or videos created by DeepFake generation methods, known as DeepFake Detection.

 Current methods focus on a variety of aspects in DeepFake detection. Some methods focused on the detection of special fingerprints during the GAN generation process \cite{yu2019attributing, marra2019gans}, or device-based fingerprint in real images (i.e., Photo response non uniformity (PRNU) ) \cite{koopman2018detection}. There are also studies that explored the differences in biological signals between real and fake videos in DeepFake detection by estimating heart rate \cite{heartrate2019, heartrate2020}. With the development of deep neural networks, multiple works leveraged different kinds of backbone models, with additional techniques such as attention mechanisms, optical flow, multi-task learning, and trained on datasets of real and DeepFake images/videos for classification \cite{wang2020cnn, deepfake_I3D, dang2020detection, amerini2019deepfake, multitask, Xception_detection}.

 Most of the learning based methods focus on the visual artifacts of the generated videos for classification, as most synthetic videos are still not perfect in visual quality. Researchers have targeted at artifacts including missing details in reflections and facial regions \cite{matern2019exploiting}, discrepancies between faces and context \cite{nirkin2020deepfake}, fake textures created in upsampling in GANs \cite{huang2020fakelocator}, inconsistencies between lip movements and audio \cite{korshunov2018deepfakes, agarwal2020detecting}, and inconsistencies in video such as lack of eye blinking \cite{deepfakeeyeblinking}, inconsistent facial landmarks between original and synthesized faces \cite{ExposingDeepFakesUsingInconsistentHeadPoses}, inconsistent emotions between audio and video \cite{Hosler_2021_CVPR}, etc. Temporal inconsistencies have also been explored for DeepFake detection on videos \cite{sabir2019recurrent, trinh2021interpretable}. However, although the above methods showed generally good performances in DeepFake detection, with rapid improvements in the output visual qualities of newer DeepFake generation methods, the artifacts-based detection methods will surely become weakened in effectiveness, and may even fail eventually. This is substantiated in the significant decrease in performance of multiple DeepFake detection methods in a recently released higher quality DeepFake dataset \cite{li2020celeb, tolosana2020deepfakes}.   

Exploring the overall behavior pattern of the talking person is vital in building a robust DeepFake detection model which can be more resistent to the rapid improvement in output qualities of newer DeepFake generation methods. However, very few methods have paid attention to this in DeepFake detection. Agarwal et al \cite{behavior1} created a dataset with the talking videos of multiple celebrities, and extracted correlation features from the facial action units (AU) which were trained with a Support Vector Machine (SVM) for DeepFake classification. Boháček and Farid \cite{protecting} similarly train an SVM on a world leader to detect deepfakes based on facial and gestural behavior. Despite a generally good classification performance, the method needs to train a separate model for each individual and the hand-crafted correlation-based features are questionable for capturing all behavioral signatures of a person such as eye gaze. Later, the authors extended their work \cite{behavior2} by using a self-supervised neural network FAb-Net \cite{FAbNet} for extracting the behavior-related features, and learned behavior and appearance-related representation with metric learning method. The model shows very good performance on several datasets, but it is unclear to interpret what behavior information is extracted from the FAb-Net. Furthermore, this method was trained and tested only on identity swap-based deepfake imagery, which are created by only replacing the face of another talking person. Differences in other visual attributes, such as hair style, face shape, etc. can still be leveraged by the model for Deepfake detection. As far as we are known of, no previous work has explicitly separated the appearance and behavioral information for studying the effect of behavioral signatures in person identification of talking videos, or has separately studied the effect of behavior pattern and utterance in the identification of fake videos. In this paper, we study both questions by keeping the appearance of the talking person but transferring the behaviors from a different person or a different utterance by using state-of-the-art lipsyncing and face reenactment models.

%% file: studydesign.tex
\section{Method}

\begin{figure*}[t]
  \centering
  \includegraphics[width=0.75\linewidth, height=2.6in]{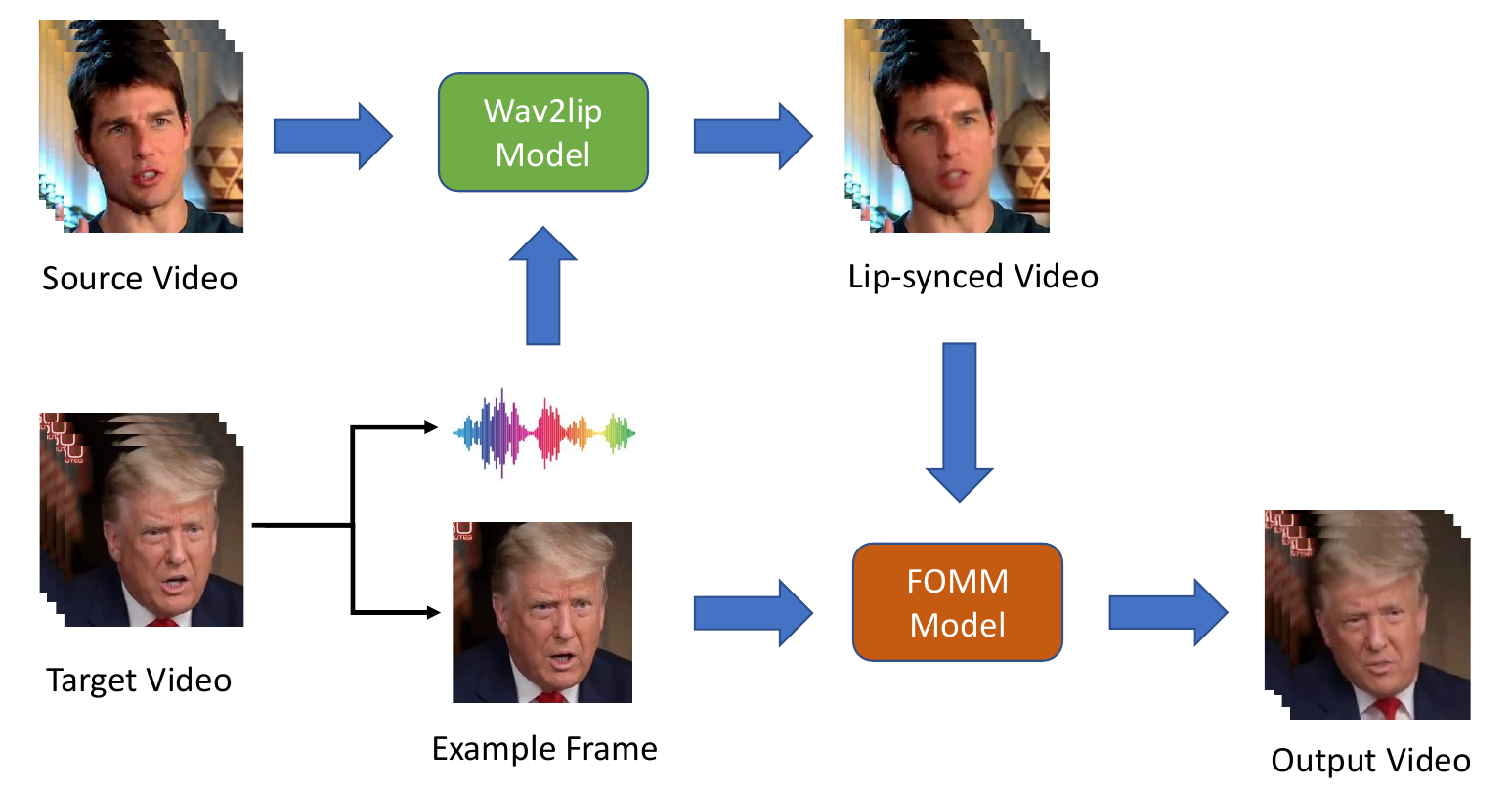}
  \caption{Overview of our method. This figure shows our video generation method in Test 1. The Wav2lip model generates lip-synced video with mouth movements corresponding to the audio of the target video. The FOMM model then uses the lip-synced video as the source video with an example target video frame for facial puppeting, to generate output video with the same person's appearance but with behavior signatures transferred from the source video. In Test 2, we removed the lower part related to FOMM by using different videos from the same person for studying the effect of utterance. In Test 3, the source video was replaced with videos acted by humans saying the same utterance to study the effect of behavior style.}
  \label{fig:overview}
\end{figure*}

Our study design involves the generation of DeepFake imagery from different sources onto a single, well known target. Our goal is to generate numerous DeepFake videos from both the original target subject, as well as from other people, then to survey a number of users about the resulting videos. In all cases, the users would see imagery of our target person, meanwhile hearing the audio of our target person. In some cases, the study participants would see clips from the original videos, in others they would see DeepFake versions using the same audio clip.

In order to minimize the potential influence of the artifacts created by the model and only investigate the effect of talking behavior and utterance, the video clip reconstructed from the model using the original video and one example frame is used as the 'real' video to be presented to the users. This allows to make a fair comparison between the real video and fake ones with behaviors transferred from another video source.

We selected Donald Trump as our target due to his popularity and distinctive talking style and behaviors. In addition, there were numerous video sources of him speaking with the camera with his head occupying the majority of the camera view. We used video clips from 2 source videos, a one-on-one interview with Trump, as well as another video from a U.S. Presidential Debate. 

We pre-processed the video clip to make the head occupy the majority of the video and located mostly at the center. To generate the video with behavior transferred from another source video, we used the Wav2lip model \cite{prajwal2020lip} for lipsyncing and First Order Motion Model (FOMM) \cite{siarohin2019first} for face reenactment. Lip imagery on every video were replaced by those from the Wav2lip model, regardless of the source, including the original video, in order to allow for consistent generation of mouth, teeth and lip appearance. 

\subsection{Technology Details}
Wav2lip \cite{prajwal2020lip} is a state-of-the-art model for generating lip-synced videos. It takes a source talking face video and an audio clip as input, and generates a lip-synced version of the input video with the mouth movements corresponding to the input audio. 
During training, it takes an audio clip and a concatenated version of input videos, which is the concatenation of a random reference segment and the ground truth segment with the lower part of the frames masked, so the ground truth segment provides the pose priors while the reference segment provides the mouth information. The input is fed to a generator which generates the output video and is trained with the L1 reconstruction loss. A pretrained lip-sync expert is used as a discriminator to penalize the incorrect lip synchronization, and a visual quality discriminator is used to improve the output quality. During inference, the input video is concatenated with the masked version of itself. 

The FOMM model \cite{siarohin2019first} is a powerful model for motion transfer and can also be used for face reenactment. It takes a pair of source and target frame as input, extracts the keypoints from both frames, and estimates the affine transformations of the source and target frames with respect to a reference frame. Then a dense motion model takes the affine transformations as input and output the optical flows from source frame to the target frame. The optical flows and the target frame are given as input to the generation module and generate the final output image with the appearance as the target and the pose as the source frame.

A comparison video between the original and the reconstructed video also shows that the FOMM provides reasonably good results by transferring facial expressions, head movements, lip and eyebrow movements to the target imagery.

\subsection{Tests}
We performed three different tests, but comparing the original actor against DeepFakes generated from different sources.

\subsubsection{Test 1}
\label{test1}
In this test, we investigated the combination effect of behavior style and utterance by using different speakers to drive the head and facial movements of our target actor.  We extracted video clips from the interviews of several other celebrities as source videos to generate DeepFake videos of Trump speaking the same words with facial movements controlled by those different source. The video generation procedure is shown in Figure \ref{fig:overview}. We selected 2 male and 2 female sources, which includes videos of Tom Cruise, Barack Obama, Taylor Swift, and Emma Watson. As the source actor is speaking a different utterance in each video, we first used the Wav2lip model to lipsync the person with audio from the original video. This lipsynced video is then used as the source video for face reenactment for transferring speaking behavior, and an example frame is selected from the original Trump interview video as the appearance reference frame. The output videos will have the appearance of Trump talking the same sentence with the behavior transferred from another celebrity, but as the transferred behavior corresponds to another person speaking a different utterance, this test will reflect on the combination effect of speaking behavior style and utterance.     

\subsubsection{Test 2}

In this test, we investigated the effect of different utterances on facial and head behavior. We extracted multiple clips from the same Trump source video and produced DeepFake videos by lip-syncing all clips talking different utterances to the audio of a selected clip. Thus, we generated DeepFake videos of which the behaviors originate from different utterances except for mouth movements which correspond to the same audio sequence. We used Wav2lip \cite{prajwal2020lip} model to lipsync other video clips to the audio of the video clip selected as "real". Similar to Test 1, the output video from the Wav2lip model using the original video and audio was chosen as the "real" video for a fair comparison. The difference between Test 1 and Test 2 is that rather than using different people speaking different utterence as sources for the DeepFake, we used the same person speaking a different utterance as the source. For this test, our goal is to have videos of different head and facial movements, but whose overall behavior style matches that of the target actor.

\subsubsection{Test 3}
In this test, we investigated the effect of speaking behavior style only. We utilized different source actors who acted out an utterance spoken by our target actor. Despite the synchronization of our source actors who intentionally matched their lips to the original audio,  we still used Wav2lip to consistent visuals on source actor's videos. We then used the head and facial movements of those sources to drive the example frame from the original Trump interview video. The DeepFake videos are created using the FOMM model, with the videos lipsynced by real persons serving as the source video for reenactment.

\begin{figure*}[t]
\centering
\begin{subfigure}{0.23\linewidth}
    \centering
    \includegraphics[width=\linewidth, height=2.0in]{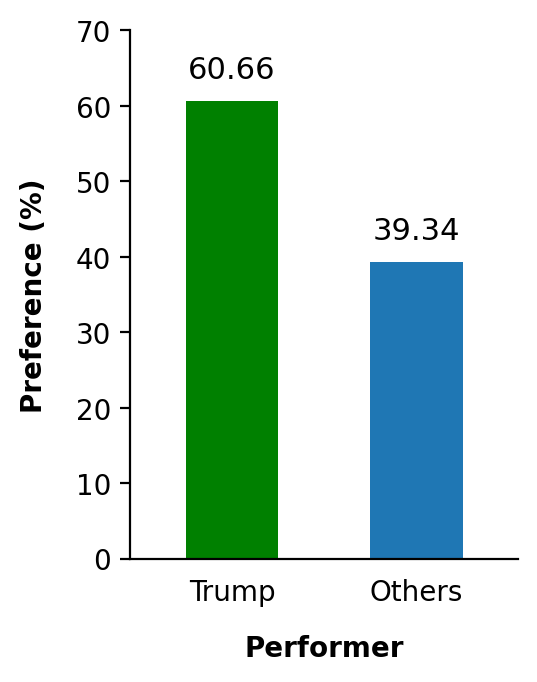}
    \caption{Preference in Naturalness}
\end{subfigure}
\begin{subfigure}{0.23\linewidth}
    \centering
    \includegraphics[width=\linewidth, height=2.0in]{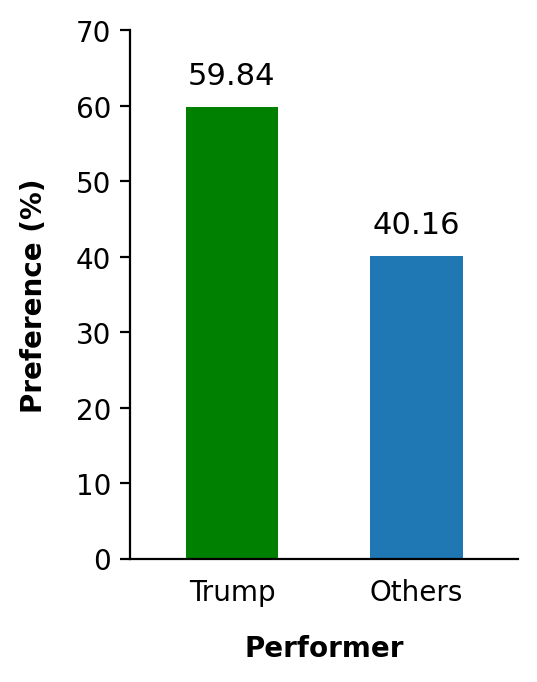}
    \caption{Preference in Engagement}
\end{subfigure}
\begin{subfigure}{0.42\linewidth}
    \centering
    \includegraphics[width=\linewidth, height=2.0in]{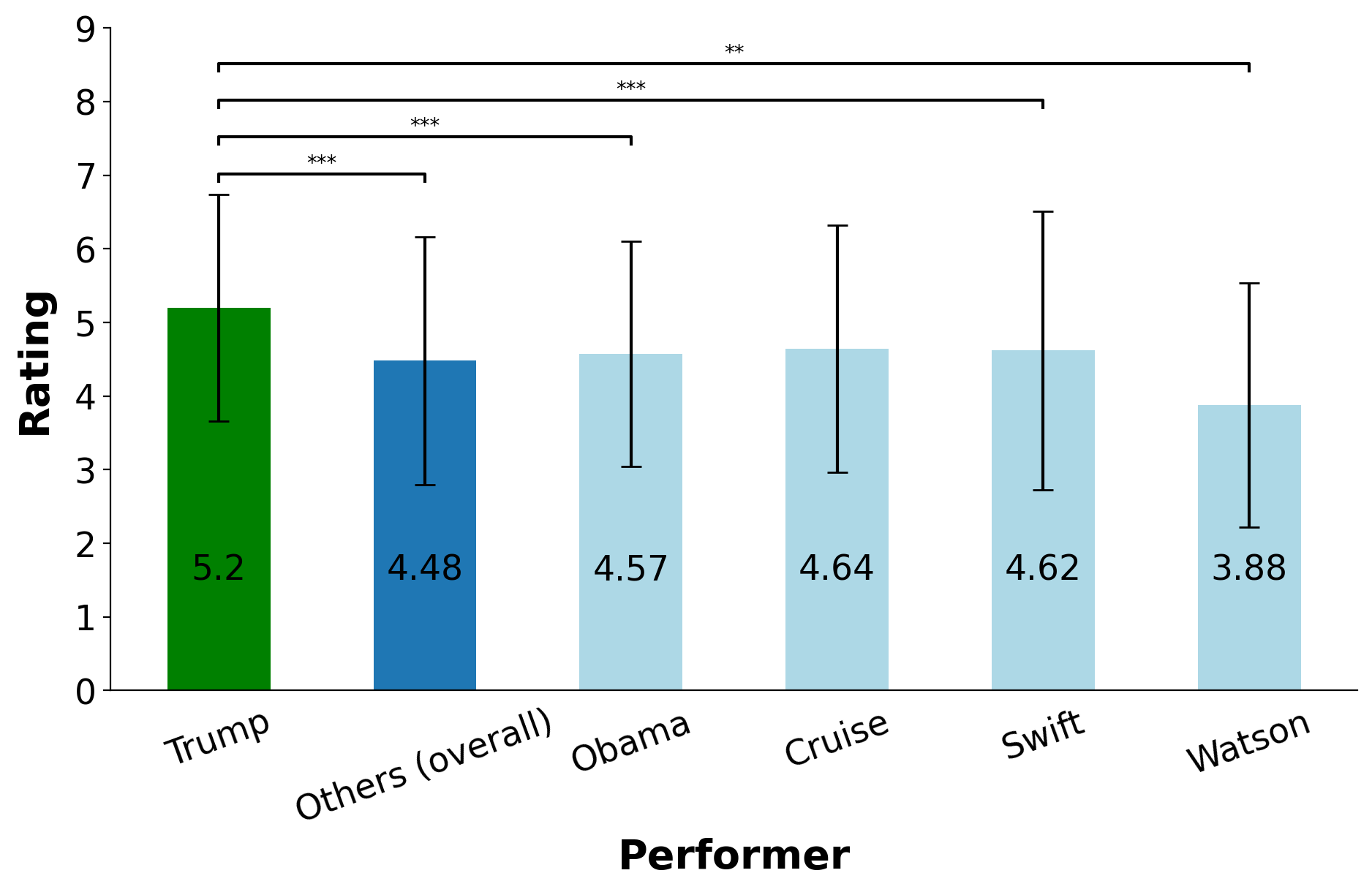}
    \caption{Individual ratings of original vs synthetic videos.}
\end{subfigure}
\caption{User Statistics of Test1. (a) and (b) show the percentage of users that showed preference for the video driven by the original actor talking video (green) or other actors' talking videos (blue). (c) shows the individual ratings of the videos from each driving source about how the person in video is like the original actor. Error bar indicates standard deviation. The number on each bar shows the average rating score. The darker blue bar shows the rating of all synthetic videos.  Statistical significance in paired t-tests is also annotated. ($^{**}p < 0.01, ^{***}p < 0.001$)}
\label{fig:test1_results}
\end{figure*}

\subsection{Study Design}

Participants were first shown a picture of the target actor and asked:
\begin{enumerate}[label=\arabic*.]
\item Do you know this person?
\item Have you seen him talking before?
\end{enumerate}

Participants who answered positively to both of those questions were then shown a series of videos. The participants view and rate the generated by videos from Tests 1, 2 and 3. In each test, the following questions are asked to the users:
\begin{enumerate}[label=\arabic*.]
\item How much is the person in the video (real/synthetic) like Trump?
Rating question: 1-7 (1: not like him, 7: exactly like him)
\item Which person in the above two videos looks more natural?
\item Which person in the above two videos are you more engaged with?
\item What movement is most important for you to get engaged with the person that you chose in the last question? Available choices: head movements,  eyebrow movements, eye movements, mouth movements, facial expressions, other.    
\end{enumerate}

There are 4 synthetic videos in Test 1, 3 in Test 2, and 2 in Test 3. For each user in each test a random synthetic video is selected and displayed with the real video. Thus, each participant viewed 6 videos; 2 from each test, and rated each one using the questions above.





%% file: results.tex
\section{Study Results}

We collected 143 responses from all participants, 122 of are identified as valid responses with users identifying themselves as familiar with Trump speaking. Participants were identified through a paid study participation service. No age or gender information was collected.

\subsection{Test 1 Results}
In Test 1 the synthetic videos are generated by videos from different person and different utterance. Figure \ref{fig:teaser} show the results as pairs of the original actor and four other actors. Figure \ref{fig:test1_results} shows the user statistics. The majority (about 60\%) of users showed a preference of the original actor video over the synthetic videos, both in naturalness and engagement. This is also reflected in the significantly higher rating scores of the original video compared to the synthetic videos in the paired t-test for all users ($t(121)= 6.29, p=5.24e^{-9}$). The individual rating scores for videos of each person as the driving source show similar results, except for Emma Watson showing a lower score compared to the average, which we speculate is due to the obvious eye looking up in the talking video. Paired t-tests on the rating scores for each individual video also showed significant higher scores for the real vs. fake videos, except for Tom Cruise.  

As shown in Figure \ref{fig:face_parts}, regarding what movement contributed most to the choice of video preference, 19 users chose 'Head' (15.32\%), 8 chose 'Eyebrow' (6.45\%), 24 chose 'Eye' (19.35\%), 40 chose 'Mouth' (32.26\%), and 33 chose 'Facial Expressions' (26.61\%). 

\subsection{Test 2}

\begin{figure*}[t]
\centering
\begin{subfigure}{0.23\linewidth}
    \centering
    \includegraphics[width=\linewidth, height=2.0in]{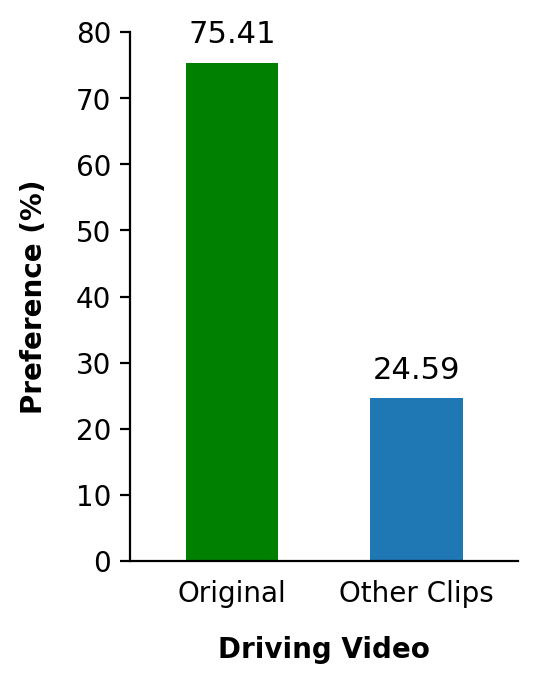}
    \caption{Preference in Naturalness}
\end{subfigure}
\begin{subfigure}{0.23\linewidth}
    \centering
    \includegraphics[width=\linewidth, height=2.0in]{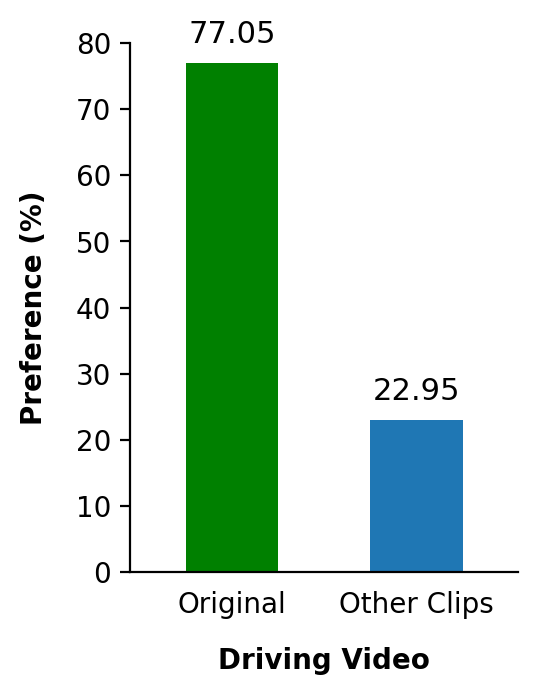}
    \caption{Preference in Engagement}
\end{subfigure}
\begin{subfigure}{0.42\linewidth}
    \centering
    \includegraphics[width=\linewidth, height=2.0in]{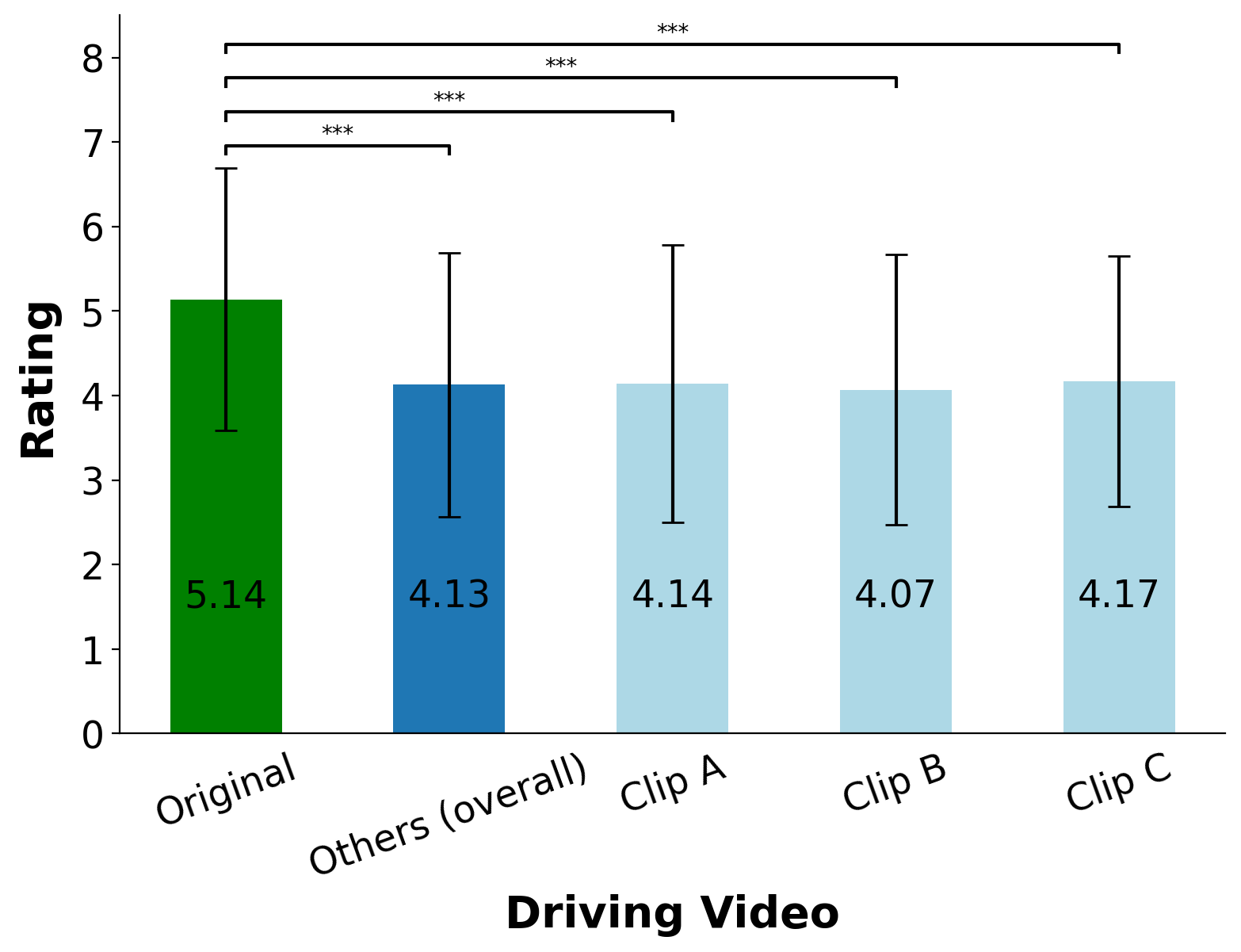}
    \caption{Individual ratings of original vs synthetic videos.}
\end{subfigure}
\caption{User Statistics of Test2. (a) and (b) show the percentage of users with a preference for the original Trump talking video (green) or the lip-synced video with a different utterance (blue). (c) shows the individual ratings of each video about how the person is like Trump. Error bar indicates standard deviation. The number on each bar shows the average rating scores. The darker blue bar shows the rating of all synthetic videos.  Statistical significance in paired t-tests is also annotated. ($^{***}p < 0.001$)}
\label{fig:test2_results}
\end{figure*}

\begin{figure*}[t]
  \centering
  \includegraphics[width=0.75\linewidth]{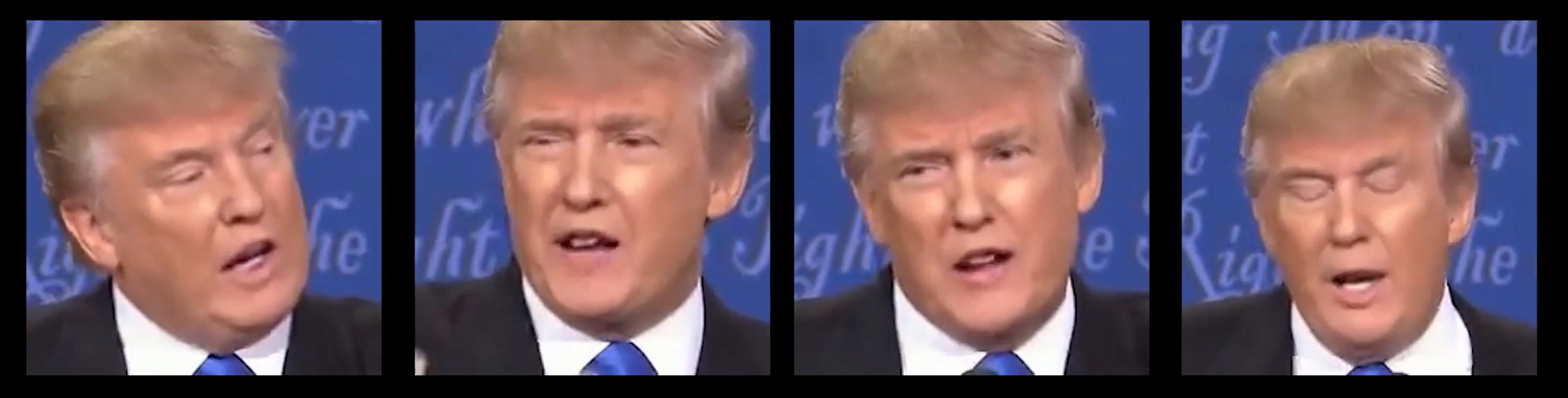}
    \includegraphics[width=0.75\linewidth]{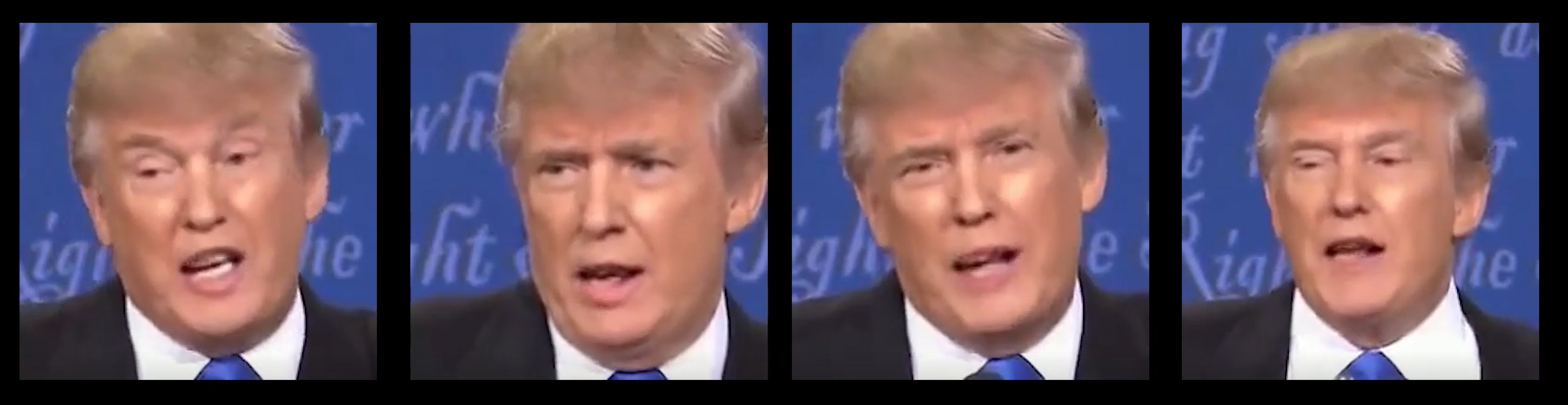}
        \includegraphics[width=0.75\linewidth]{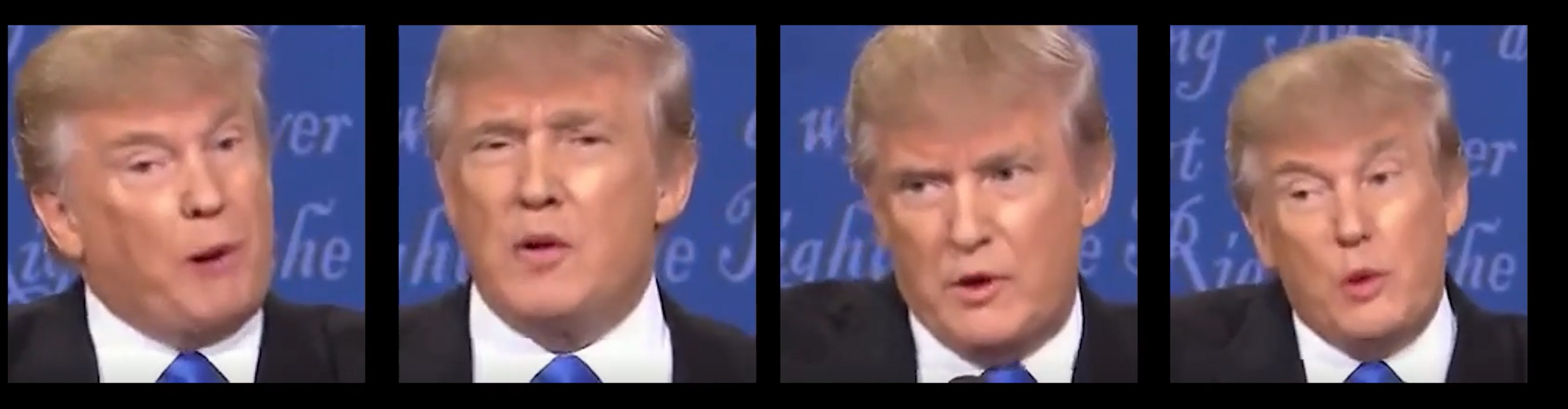}
  \caption{Frame examples from Test 2. All videos play the same audio and have matching mouth movements to audio. Leftmost video is the original video after modifying with lip sync process. The three other videos on the right are taken from a moment during of the same actor, and thus have head and facial behaviors that of a different utterance.}
  \label{fig:test2example}
\end{figure*}

In Test 2 the synthetic videos are generated by videos from the same person with a different utterance. Examples of the output are shown in Figure \ref{fig:test2example}. Figure \ref{fig:test2_results} shows the user statistics. A larger portion of the users showed a preference of the original actor video over the lipsynced videos, both in naturalness and engagement. Paired t-test also shows a significantly higher rating scores of the original video compared to the synthetic videos ($t(121)= 7.73, p=3.5e^{-12}$). The individual rating scores for videos of each lipsynced video have almost no difference with the overall average score. Paired t-tests on the rating scores for each individual video also showed significant higher scores for the real vs. fake videos.

As shown in Figure \ref{fig:face_parts}, regarding what movement contributed most to the choice of video preference, 30 users chose 'Head' (24.19\%), 17 chose 'Eyebrow' (13.71\%), 25 chose 'Eye' (20.16\%), 29 chose 'Mouth' (23.39\%), and 23 chose 'Facial Expressions' (18.55\%).

\subsection{Test 3}
\begin{figure*}[t]
\centering
\begin{subfigure}{0.23\linewidth}
    \centering
    \includegraphics[width=\linewidth, height=2.0in]{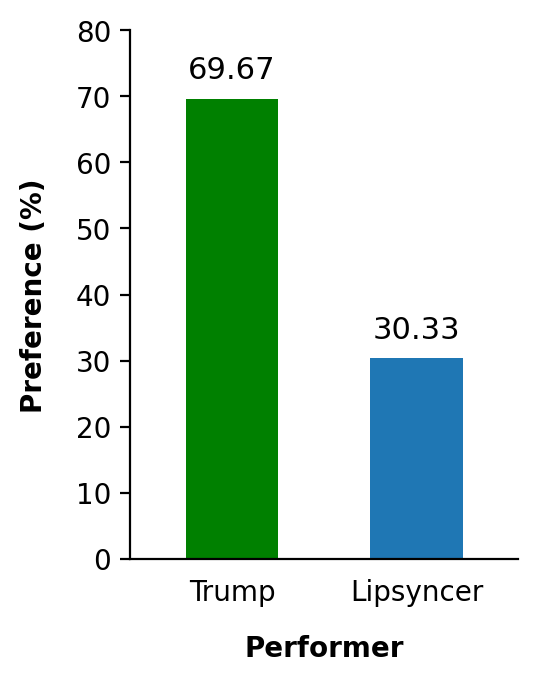}
    \caption{Preference in Naturalness}
\end{subfigure}
\begin{subfigure}{0.23\linewidth}
    \centering
    \includegraphics[width=\linewidth, height=2.0in]{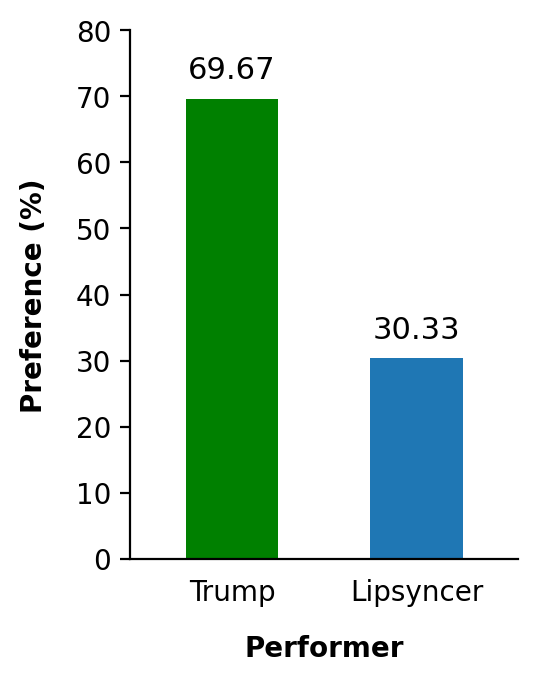}
    \caption{Preference in Engagement}
\end{subfigure}
\begin{subfigure}{0.42\linewidth}
    \centering
    \includegraphics[width=\linewidth, height=2.0in]{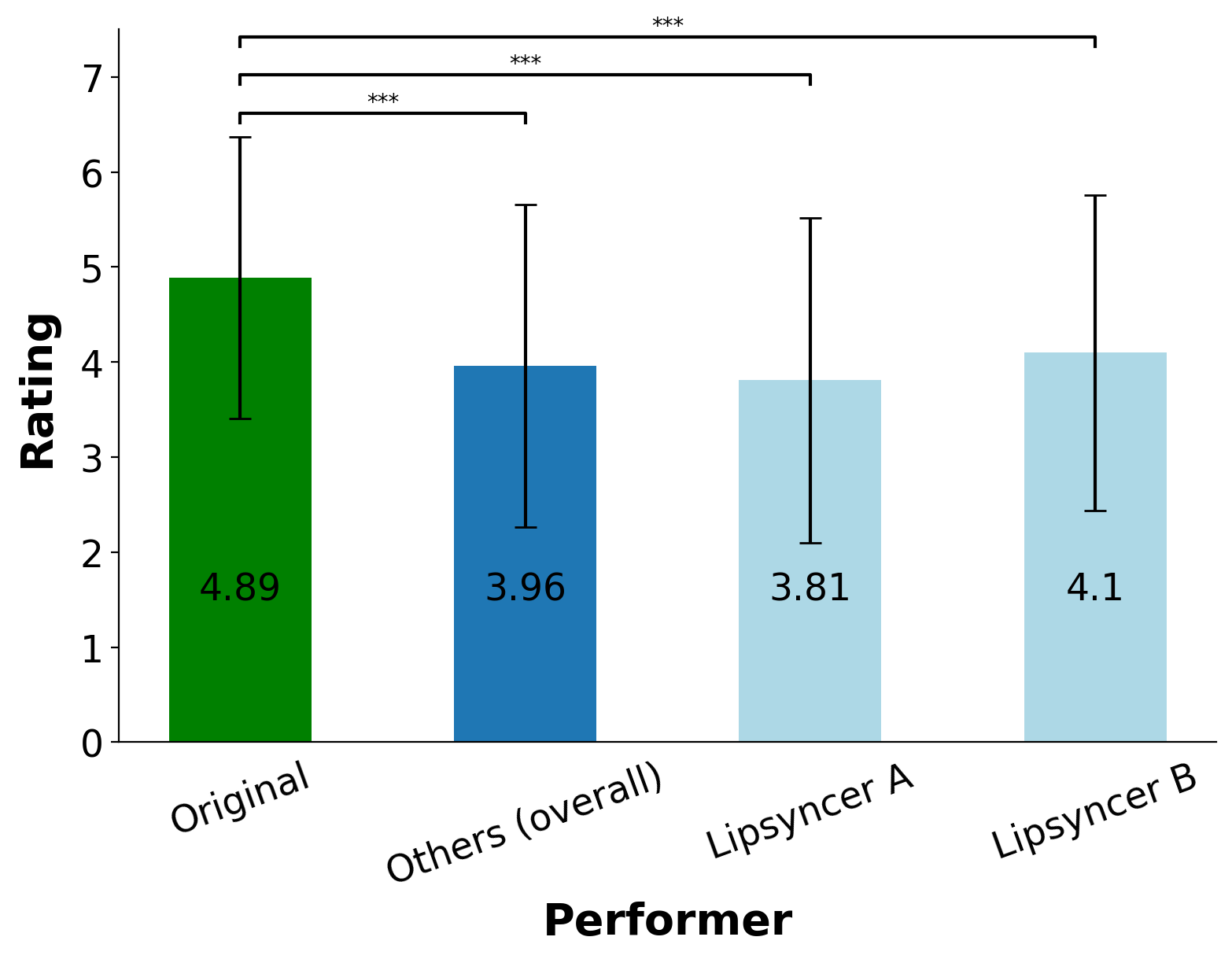}
    \caption{Individual ratings of original vs synthetic videos.}
\end{subfigure}
\caption{User Statistics of Test3. (a) and (b) show the percentage of users with a preference for videos driven by the original Trump talking video (green) or videos from other lipsyncers (blue). (c) shows the individual ratings about how the person in video is like Trump. Error bar indicates standard deviation. The number on each bar shows the average rating score. The darker blue bar shows the rating of all synthetic videos. Statistical significance in paired t-tests is also annotated. ($^{***}p < 0.001$)}
\label{fig:test3_results}
\end{figure*}

In Test 3 the synthetic videos are generated by videos from the different actors using the same utterance and audio. Figure \ref{fig:test3_results} show the user statistics. The original video still shows much larger portion of preference, with about 70\% of users preferring the original video, both in naturalness and engagement. Paired t-test also shows a significantly higher rating scores of the original video compared to the synthetic videos ($t(121)= 6.11, p=1.26e^{-8}$). Figure \ref{fig:test3_results} shows a small variation of the rating scores between different lipsyncers. Paired t-tests on the rating scores also showed significant higher scores for the real vs. fake videos for videos from both lipsyncers.

\begin{figure}
\centering 
\includegraphics[width=\linewidth]{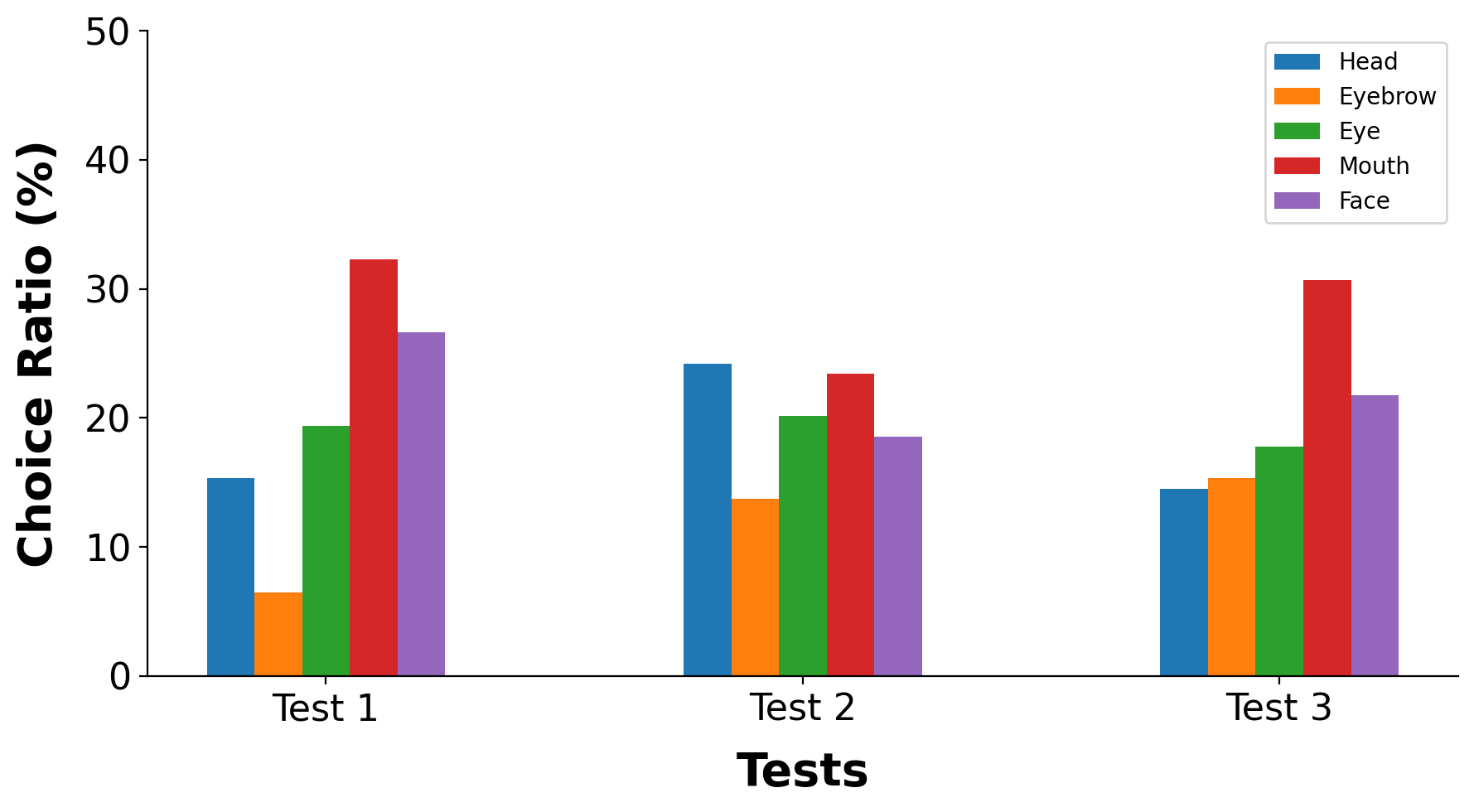}
\caption{Importance of specific movements. The most important movements selected by the users for making the choices in all 3 tests are visualized. }
\label{fig:face_parts}
\end{figure}

As shown in Figure \ref{fig:face_parts}, regarding what movement contributed most to the choice of video preference, 18 users chose 'Head' (14.52\%), 19 chose 'Eyebrow' (15.32\%), 22 chose 'Eye' (17.74\%), 38 chose 'Mouth' (30.65\%), and 27 chose 'Facial Expressions' (21.77\%).

%% file: discussion.tex
\section{Discussion}
Our tests attempt to explore the effect on differing sources on a resulting synthetic video, including the appropriateness of nonverbal behavior, as well as the source actor. In Test 1, we created synthetic videos using the same audio from different utterances from different people. In Test 2, we created synthetic videos using the same audio from different utterances from the same person. In Test 3, we created synthetic videos using the same audio from different people. Because the quality of the synthetic video is potentially lower than that of the source video, we applied a similar process to the original video in order to equalize the quality of the resulting videos, thus making sure the use of Wav2lip/FOMM can alter the original video in the same way it is used to alter the resulting videos regarding potential artifacts. 

In all tests, a significantly larger portion of the users preferred the reconstructed original video compared to the synthetic videos. Paired t-test on the users' ranking scores of how much the person is like the original actor also showed significantly higher scores for the original videos compared to the synthetic ones. Based on the results of all 3 user tests, we can draw the following conclusions:

Our study indicates that for a person with a distinctive speaking style:
\begin{itemize}
    \item It can be identified by humans if his/her talking behavior is replaced by that of another person
    saying a different utterance.
    \item It can be identified by humans if his/her talking behavior is replaced by himself/herself saying a
    different utterance.
    \item It can be identified by humans if his/her talking behavior is replaced by that of another person saying the same utterance.
\end{itemize}

Our study indicates that both the speaking behavior style and the correspondence between speaking behavior and utterance play vital roles in the
non-appearance aspect in the identification of a person. This provides evidence that the distinct speaking style of a person and the correspondence between speaking behavior and utterance can serve as important clues for DeepFake detection even for synthetic video with perfect visual appearance in the future.

The users' choices of which part of the facial movement contributed most to their preference in all 3 tests are summarized in Figure \ref{fig:face_parts}. In Test 1 and Test 3, both the mouth movements and facial expressions are the two dominant attributes that are chosen by the users. The common point between Test 1 and Test 3 is that they are both from different persons, so it may lead to an inference that the differences in speaking styles between persons reflects most obviously in mouth movements and facial expressions. However it still needs further investigation as in both tests the mouth movements are already replaced by the wav2lip generated mouth movements. Nevertheless, we can still argue that Wav2lip captures the lip movement patterns in the mouth as in its design it takes in the mouth movement patterns from the reference segment.
In Test 2, we can see that the head movement shows a much higher percentage compared to Test 1 and Test 3, so it may suggest that head movement has a stronger effect when there are differences in utterances.

Therefore, from the above discussion, we can make another weak conclusion that mouth movement and facial expression tend to play more important roles in distinguishing the speaking behavior styles of different persons, while head movement tend to have a greater importance in sensing the correspondences between talking behaviors and utterances. This conclusion may need further investigation and can be addressed more specifically in future studies.

\section{Conclusion}
Many DeepFake detection methods rely on the visual quality of the resulting imagery to determine the presence of an authentic or synthetic video. With the improvement in image generation methods, we expect to see in the near future synthetic videos that evade these methods. In this work, we explore whether a person has their own behavioral signature that is recognizable by others who are familiar with that person, and whether that signal can be used to distinguish a synthetic video from a real one. The results of our study show that both behavioral signature and correspondences with utterance can significantly affect humans' judgements of the naturalness of a video. This provides evidence of the necessity for leveraging behavioral signature and the correspondence between behavior and utterance in Deepfake Detection, which are overlooked by models that examine visual quality alone.